\documentclass[letterpaper, 10 pt, conference]{ieeeconf}  

\IEEEoverridecommandlockouts                              

\usepackage{graphics} 
\usepackage{epsfig} 
\usepackage{mathptmx} 
\usepackage{times} 
\usepackage{amsmath} 
\usepackage{amssymb}  

\title{\LARGE \bf
Ditto in the House: Building Articulation Models of Indoor Scenes through Interactive Perception
}

\author{
    Cheng-Chun Hsu$^{1}$ and Zhenyu Jiang$^{1}$ and Yuke Zhu$^{1}$
    \thanks{
        $^{1}$ Department of Computer Science, the University of Texas at Austin. Correspondance to {\tt\small chengchun@utexas.edu}
    }
}

\usepackage{xspace}
\usepackage{graphicx}
\usepackage{amsmath}
\usepackage{amssymb}
\usepackage{booktabs}
\usepackage{multirow}
\usepackage{cellspace, tabularx}
\usepackage{xcolor}
\usepackage{bbm}
\usepackage{csquotes}
\usepackage{hyperref}
\usepackage{algorithm}
\usepackage{algpseudocode}
\algrenewcommand\algorithmicrequire{\textbf{Input:}}
\algrenewcommand\algorithmicensure{\textbf{Output:}}
\usepackage{gensymb}
\usepackage[english]{babel} 
\usepackage{blindtext}
\DeclareUnicodeCharacter{0301}{'}

\usepackage[backend=biber,
            hyperref=true,
            url=false,
            isbn=false,
            doi=false,
            backref=false,
            style=ieee,
            citestyle=numeric-comp,
            sorting=nyt,
            block=none]{biblatex}
\usepackage[font=footnotesize]{caption}
\renewcommand{\bibfont}{\small}
\makeatletter
\DeclareRobustCommand\onedot{\futurelet\@let@token\@onedot}
\def\@onedot{\ifx\@let@token.\else.\null\fi\xspace}

\def\eg{\emph{e.g}\onedot} 
\def\ie{\emph{i.e}\onedot}

\def\etal{\emph{et al}\onedot}
\makeatother

\begin{document}

\maketitle
\thispagestyle{empty}
\pagestyle{empty}


\begin{abstract}
Virtualizing the physical world into virtual models has been a critical technique for robot navigation and planning in the real world. To foster manipulation with articulated objects in everyday life, 
this work explores building articulation models of indoor scenes through a robot's purposeful interactions in these scenes. 
Prior work on articulation reasoning primarily focuses on siloed objects of limited categories. To extend to room-scale environments, the robot has to efficiently and effectively explore a large-scale 3D space, locate articulated objects, and infer their articulations. We introduce an interactive perception approach to this task. Our approach, named \textit{Ditto in the House}, discovers possible articulated objects through affordance prediction, interacts with these objects to produce articulated motions, and infers the articulation properties from the visual observations before and after each interaction. It tightly couples affordance prediction and articulation inference to improve both tasks. We demonstrate the effectiveness of our approach in both simulation and real-world scenes. Code and additional results are available at \href{https://ut-austin-rpl.github.io/HouseDitto/}{\url{https://ut-austin-rpl.github.io/HouseDitto/}}

\end{abstract}
\section{Introduction}
Virtualizing the real world into virtual models is a crucial step for robots to operate in everyday environments. Intelligent robots rely on these models to understand the surroundings and plan their actions in unstructured scenes. 
%
Recent advances in structure sensors and SLAM algorithms~\cite{newcombe2011kinectfusion,newcombe2015dynamicfusion,geiger2011stereoscan} have offered ways to construct static replicas of real-world scenes at unprecedented fidelity. 
These static replicas facilitate mobile robots to localize themselves and navigate around. Nevertheless, real-world manipulation would require a robot to depart from reconstructing a static scene to unraveling the physical properties of objects. In particular, to physically interact with articulation objects, such as cabinets and doors, commonly seen in daily environments, the robot needs to understand their kinematics and articulation properties. Motivated by this goal, this work aims to enable robots to automatically build articulated 3D models of indoor scenes from their purposeful interactions.

Understanding object articulation has been a long-standing challenge in computer vision and robotics. In recent years, a series of data-driven approaches have attempted to infer articulation from static visual observations by training models on large 3D datasets~\cite{li2020category,weng2021captra}. These models rely on category-level priors, limiting their generalizations to a handful of preset categories. A parallel thread of research~\cite{hausman2015active,martin2014online} uses interactive perception~\cite{bohg2017interactive}. They leverage physical interaction to emit visual observations of articulated motions from which they estimate the articulation properties. These works primarily focus on individual objects, whereas scaling to room-sized environments requires the robot to efficiently and effectively explore the large-scale 3D space for meaningful interactions.

\begin{figure}[t]
    \centering
    \includegraphics[width=\linewidth]{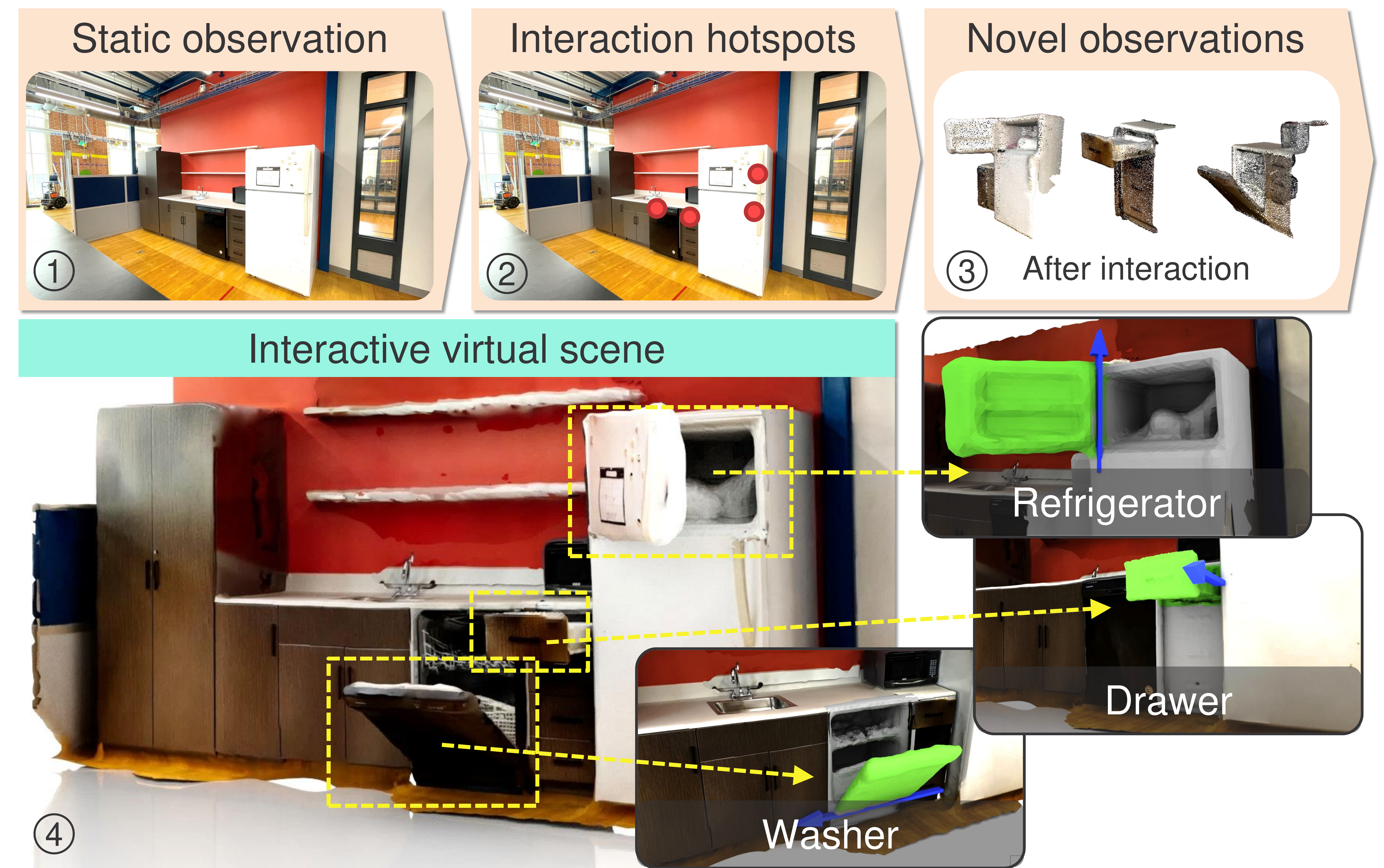}
    \vspace{-5mm}
    \caption{
        \textbf{Building scene-level articulation models through interactive perception.} 
        From an initial observation of an indoor scene, our approach infers the interaction hotspots, guiding the robot to interact with articulated objects. After that, the robot collects the observations before and after the interactions. Based on the observed articulated motions, it builds the articulation models of individual objects and aggregates them into a scene-level articulation model.
    }
    \vspace{-2mm}
    \label{fig: teaser}
\end{figure}

We introduce \textbf{Ditto in the House}, an approach to building the articulation model of an indoor environment through a robot's self-directed exploration. 
The robot discovers and physically interacts with the articulated objects in the environment. Based on the visual observations before and after the interactions, the robot infers the articulation properties of the interacted objects.
Our approach requires the robot to identify regions of possible articulations in a large-scale 3D space, manipulate the objects to create articulated motions, and infer the underlying kinematic parameters from partial observations. 

The foremost challenge is determining the most effective actions for probing the environment, \textit{i.e.}, those most likely to discover articulations. We cast this problem as inferring scene affordance from visual observations~\cite{mo2021where2act,nagarajan2020learning,do2018affordancenet}. Specifically, we train a model to predict affordance based on the robot's past interaction experiences with a large collection of procedurally generated 3D scenes in simulation.
Furthermore, we introduce an iterative refinement procedure that uses the initial observations of articulated motions as visual cues to refine the affordance predictions. This refined affordance, in turn, guided the robot's subsequent interactions to obtain more informative observations for articulation inference.

Figure~\ref{fig: teaser} illustrates the high-level steps of our approach. It discovers the interaction hotspots (\emph{i.e.}, promising locations for probing actions), creates observations of articulated motions through action, and infers articulation from the observations. 
To locate interaction hotspots, we use an affordance model to predict an affordance map from the point cloud observation of an indoor scene.
At the location of each interaction hotspot, the robot tries interacting with the objects and collects egocentric observations of the object before and after its motion induced by the robot's actions. Based on our prior work Ditto~\cite{jiang2022ditto}, we design an articulation estimation model to infer the object articulation from these observations. The articulation estimation model takes the robot's action and visual observations as input and predicts the articulation parameters of the objects.
We tightly couple the affordance prediction and articulation estimation in an iterative process. We update the affordance prediction based on the initial estimation of mobile parts of the articulated object since these parts indicate actionable regions. The robot interacts with the object following the updated affordance, creating more visible articulation motions that facilitate articulation inference.

We evaluate our approach on the CubiCasa5K~\cite{kalervo2019cubicasa5k} dataset with the iGibson~\cite{li2021igibson} simulator. Quantitative results demonstrate that our approach successfully discovered around 40\% more parts than our baseline with higher precision. Our approach also gives a 45\% absolute segmentation IoU boost compared with the baseline. Moreover, the ablation studies confirm that the iterative refinement process benefits both affordance prediction and articulation model estimation. Finally, we apply our approach to a real-world kitchen scene and successfully build an articulation model of this environment.

\section{Related work}

\subsection{Visual Affordance Prediction}
Predicting affordance~\cite{james1979ecological} from visual observations is an essential ability for robots to plan their actions for interacting with real-world objects. A series of research learns visual affordance prediction from human videos~\cite{brahmbhatt2019contactdb,nagarajan2019grounded,hamer2010object,fouhey2012people}. These videos reveal strong clues about how humans interact with their environment. However, affordances learned from these videos are often specific to human morphologies and fall short of generalizing to robot hardware. Another line of research acquires active training data with simulated or real robots for affordance learning~\cite{mo2021where2act,khansari2020action,jiang2021synergies,zeng2018learning,nagarajan2020learning}. In these methods, the robots explore diverse interactions with objects and scenes and learn affordances from their embodied experiences. Following this self-exploratory learning paradigm, 
we collect exploration data in simulation and train a visual affordance prediction model. This model guides our agent to discover articulated objects in the scene and interact with them for articulation inference.

\subsection{Articulation Model Estimation}

Articulation models represent the object parts and the kinematics (and sometimes dynamics) relationships between them. Earlier works model the object partonomy and articulation with probabilistic methods~\cite{dearden2005learning,sturm2008adaptive,sturm2008unsupervised,sturm2009learning}. They typically rely on markers or handcrafted features to track the mobile parts, limiting their practical applicability in natural environments. In recent years, deep learning methods~\cite{omran2018neural,li2020category,wang2019shape2motion,weng2021captra,noguchi2021neural,hameed2022learning} have been employed for articulation estimation from raw sensory data. The majority of these works predict articulation models from single observations. They rely on category-level prior to estimating articulation parameters. For this reason, they are category-dependent and cannot generalize to diverse daily objects in indoor scenes. 
Another line of work leverages physical interaction to create novel sensory stimuli and infer the articulation model based on object state changes. Katz \etal~\cite{katz2013interactive} is the pioneer work that introduces interactive perception~\cite{bohg2017interactive} for articulation model estimation. Some following works extend it with hierarchical recursive Bayesian filters~\cite{martin2016integrated}, probabilistic models~\cite{sturm2011probabilistic}, and feature tracking~\cite{pillai2015learning}. Recent work in this direction utilizes the informative motion created by handcrafted strategies~\cite{jiang2022ditto} or learned policy~\cite{kumar2019estimating,gadre2021act,lv2022sagci} and estimate articulation models with geometric deep learning techniques. The methods above are designed for individual objects. They cannot be easily extended to large indoor environments with multiple objects at unknown locations. In contrast, our approach develops a category-independent approach that actively explores the environment and builds scene-level articulation models.

\begin{figure*}[t]
    \centering
    \includegraphics[width=\textwidth]{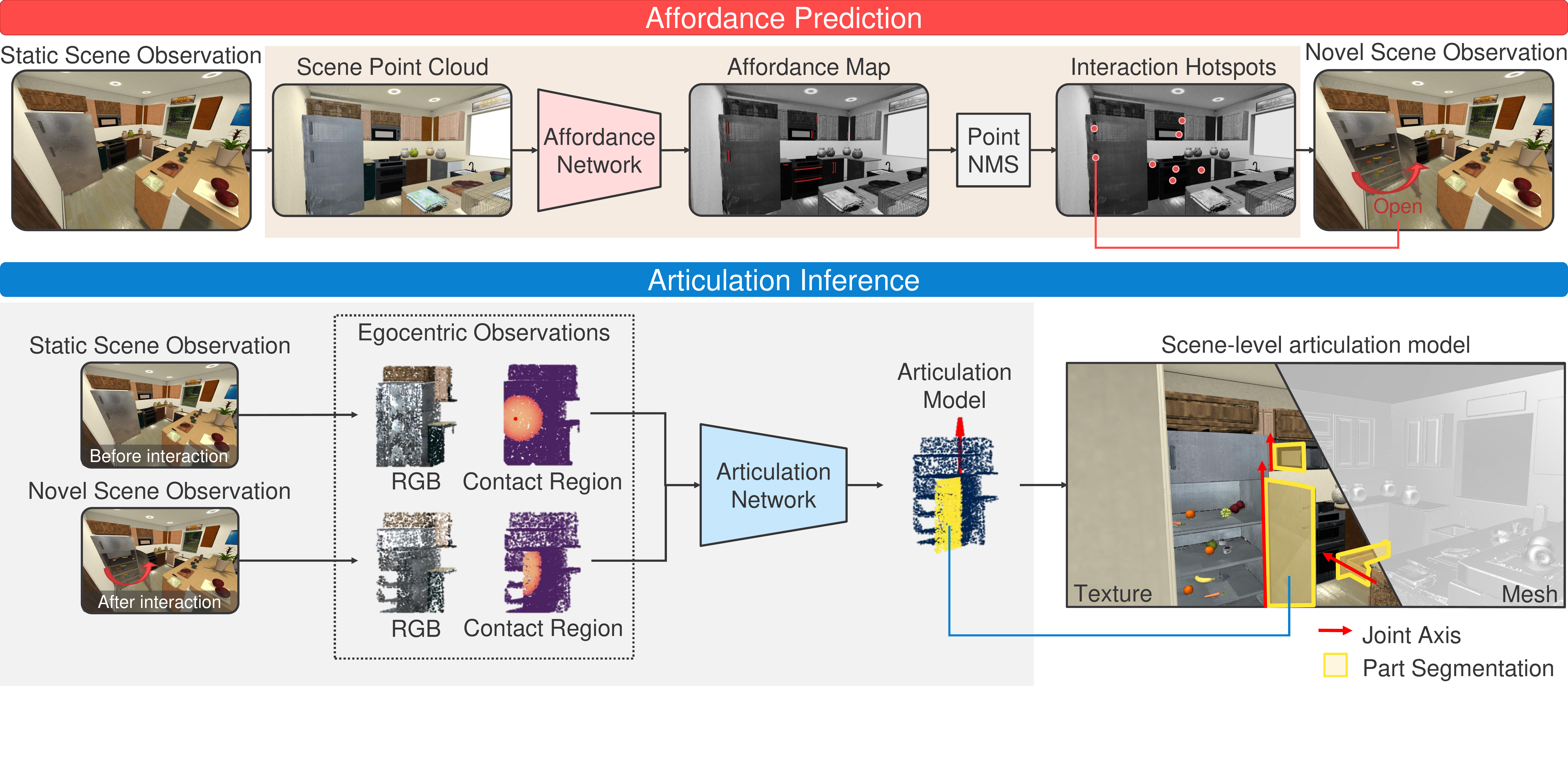}
    \vspace{-17mm}
    \caption{
        \textbf{Overview of model components.} 
        Our approach consists of two stages --- affordance prediction and articulation inference. During affordance prediction, we pass the static scene point cloud into the affordance network and predict the scene-level affordance map. By applying point non-maximum suppression (NMS), we extract the interaction hotspots from the affordance map. Then, the robot interacts with the object based on those contact points. During articulation inference, we feed the point cloud observations before and after each interaction into the articulation model network to obtain articulation estimation. By aggregating the estimated articulation models, we build the articulation models of the entire scene.
    }
    \label{fig: framework}
\end{figure*}

\section{Problem Formulation}
We explore the problem of building an articulation model of an indoor scene populated with articulated objects. An articulated object consists of multiple parts, and their connecting joints constrain the relative motion between each pair of parts. We assume there exists an initial point cloud observation $P_s \in \mathbb{R}^{N_s \times 6}$ of the scene, where each point is a vector of its $(x, y, z)$ coordinate and $(R, G, B)$ color. $N_s$ is the number of points. This work aims to segment the parts and estimate the articulation models $M$ of the scene through interactive perception. Our approach actively interacts with the objects to estimate their articulation models.

  For each object part, we identify the interaction hotspots~\cite{nagarajan2019grounded}, \textit{i.e.}, the locations where the robot can successfully manipulate the articulated objects, and infer the articulation model. The articulation model represents the kinematic constraints between object parts. We parametrize articulation following ANCSH~\cite{li2020category}. Prismatic joint parameters are the direction of the translation axis $u^p \in \mathbb{R}^{3}$ and the joint state $s^p \in \mathbb{R}$. Revolute joint parameters are the direction of the revolute axis $u^r \in \mathbb{R}^{3}$, a pivot point $q \in \mathbb{R}^{3}$ on the revolute axis, and the joint state $s^r \in \mathbb{R}$. The joint state $s^p$ and $s^r$ are the relative changes of the joint state before and after the interaction. 

The process starts with finding interaction hotspots $H_s \in \mathbb{R}^{N_h \times 3}$ from an initial scene observation $P_s$, where $N_h$ denotes the number of interaction hotspots in the scene. From the point cloud observation $P_s$, an affordance prediction model predicts an affordance map $A_s \in \mathbb{R}^{N_s}$ and samples its peak locations as interaction hotspots. 

At each interaction, the robot applies forces to the object hotspot to produce potential articulated motions. The robot captures two egocentric partial point clouds $P \in \mathbb{R}^{N_o \times 3}$ and $P' \in \mathbb{R}^{N_o' \times 3}$ that center on the interaction hotspot before and after the interaction, where $N_o$ and $N_o'$ denotes the number of points on object point clouds. The robot also records the corresponding contact locations $c, c' \in \mathbb{R}^{3}$. Given these object point clouds $P, P'$ and contact locations $c, c'$, an articulation inference model segments the point cloud into static and mobile parts and estimates a set of articulation parameters,~\ie, $\{u^p, s^p\}$ (prismatic) or $\{u^r, q, s^r\}$ (revolute), of the joints connecting two object parts. Finally, we map the estimated articulation model of each object from the global frame. The set of these articulation models constitutes the scene-level articulation model $M$.

\section{Approach}
We now present our approach to building the articulation model of indoor scenes, as illustrated in Figure~\ref{fig: framework}. In the following subsections, we introduce the three key components of our approach, affordance prediction, articulation estimation, and iterative refinement of both affordance and articulation.

\subsection{Affordance Prediction}
At the initial stage, we must identify potential interaction regions and discover the articulated objects. We leverage PointNet++~\cite{qi2017pointnet++} to estimate scene affordance from an initial point cloud observation of the scene. Formulating affordance prediction as a point-wise binary classification problem, we feed the scene point cloud into PointNet++ and obtain a point-wise affordance map. Based on the map, the robot actively explores the scene and interacts with objects at the corresponding locations.
Since dense affordance prediction may introduce too many potential locations to probe, we apply non-maximum suppression (NMS)~\cite{girshick2014rich} to extract peaks from each region as the \textit{interaction hotspots}.

Our approach works as follows: first, we select the point with the maximum score and add it to the preserving set. Then we suppress its neighbors by a certain distance threshold. We repeat this process until all points are added to the preserving set or suppressed. The points left in the preserving set are the interaction hotspots. The robot physically interacts with objects at the corresponding location for each interaction hotspot. These interactions create informative motions for further articulation inference.

\subsection{Articulation Inference}
\label{sec: articulation}
After each effective interaction, we observe changes in the articulated configurations of the object. The robot collects egocentric observations of the object before and after the interactions. We develop an articulation network to infer the articulation model from these observations.
%






The network is built on top of our prior work Ditto~\cite{jiang2022ditto}. 
Given a pair of visual observations of an object before and after its articulated motion, Ditto simultaneously predicts the 3D geometry and articulation model of the object. Original Ditto takes the observations as input and does not know the actions that create the motion in the observation. In contrast, our agent physically interacts with the object and then collects the observations. So we incorporate the knowledge of the interaction and use the contact regions of the interaction as part of the input to our network.
For each interaction pair, the robot captures the observation point clouds $P$ and $P'$ by interacting at the corresponding contact locations $c$ and $c'$. We create Gaussian heatmaps centered around the $c$ and $c'$ over the point cloud $P$ and $P'$, respectively. These heatmaps represent the regions where interactions occur. The locations of contact regions during interaction reveal a vital clue about the mobile part region and the underlying kinematic constraint. Therefore, we feed both the point cloud observations and the heatmaps of contact regions into the network for articulation inference.

Finally, the network outputs the articulation joint parameters, joint state, and part segmentation of each articulated object. To build the final articulation models, we aggregate the estimated object-level articulation models into a scene-level model.

\subsection{Iterative Refinement of Affordance and Articulation}
The inference of the articulation model primarily relies on the visual observations captured during the interactions. The estimated articulation model could have a higher accuracy if the observations covered significant articulation motions and a complete view of the object's interior. However, these observations may be partially occluded due to ineffective actions, results from imprecise affordance predictions and manipulation failures. Partially occluded observations lead to inaccurate articulated predictions. Empirically, we find that articulation estimation of a fully opened revolute joint,~\eg, $> 30\degree$, is more accurate in terms of angle error than one with an ajar joint. 

In practice, we find that even a coarsely estimated articulation model could reveal useful clues about part-level affordance. The articulation model provides the estimated location of the joint axis and part segmentation, which we can use to predict the possible kinematic motions of the object parts. As we want to produce a larger motion of the object part, we update the affordance based on the motion predicted from the articulation model. Accordingly, we develop an iterative procedure of interacting with partially opened joints and refining the articulated predictions.

\begin{figure}[t]
    \centering
    \includegraphics[width=\linewidth]{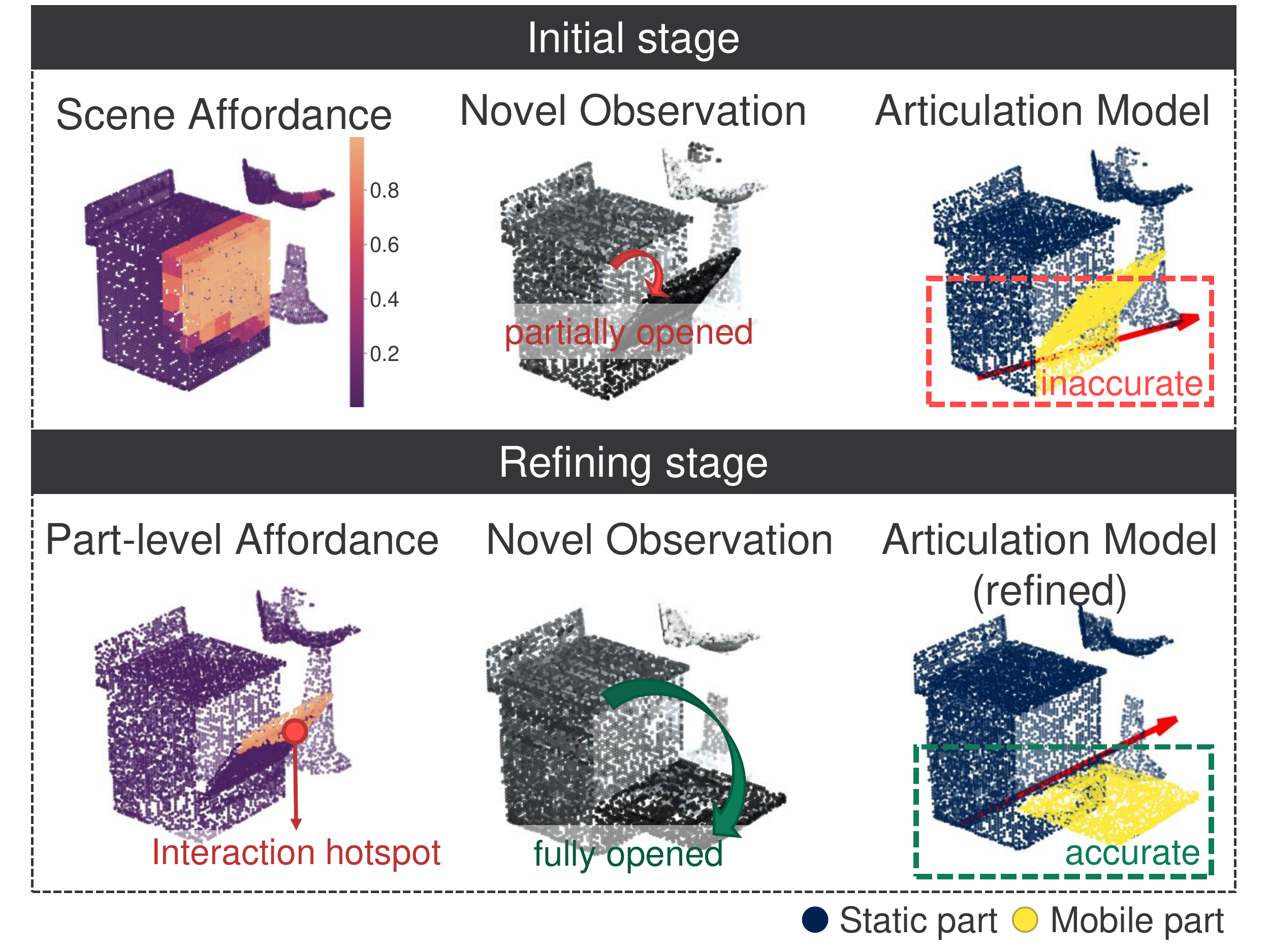}
    \vspace{-5mm}
    \caption{
        \textbf{Iterative refinement of affordance and articulation.} 
        In the initial stage, the object is partially opened due to the imprecise affordance prediction, which results in an inaccurate articulation estimation. In the refining stage, we refine object affordance based on the previous articulation estimation. The consequent new interaction fully opens the object and reveals the interior surface, thus improving the articulation estimation.
    }
    \label{fig: synergies}
    \vspace{3mm}
\end{figure}

As shown in Figure~\ref{fig: synergies}, the estimated affordance at the initial stage did not consider the articulation model and spread uniformly over the surface of the mobile part. The extracted interaction hotspot can be close to the axis. This location and the hand-crafted action primitive produce less torque on the revolute axis. As a result, the robot fails to fully open the revolute joint. In our experiments, only $10\%$ of the revolute joints are opened up to $> 30\degree$ during the initial interaction, which hinders the performance of articulation estimation.

To improve the articulation model inferred at the first stage, we exploit the potential motion information from the previous articulation model and extract the object-level affordance. We refine the affordance prediction by selecting a pair of locations and actions to produce the most significant articulation motion. Given the joint axis and part segment, we select the point in the predicted mobile part farthest from the joint axis as our next interaction hotspot. We set the force direction as the moment of the axis, \ie, the cross-product between the axis direction and the projection from the interaction hotspot to the axis. Finally, we interact with the object following the updated interaction hotspots and force directions and collect new observations. The new observations are less occluded and result in an accurate articulation estimation.

\section{Experiments}


\begin{figure}[t]
    \centering
	\includegraphics[width=1.0\linewidth]{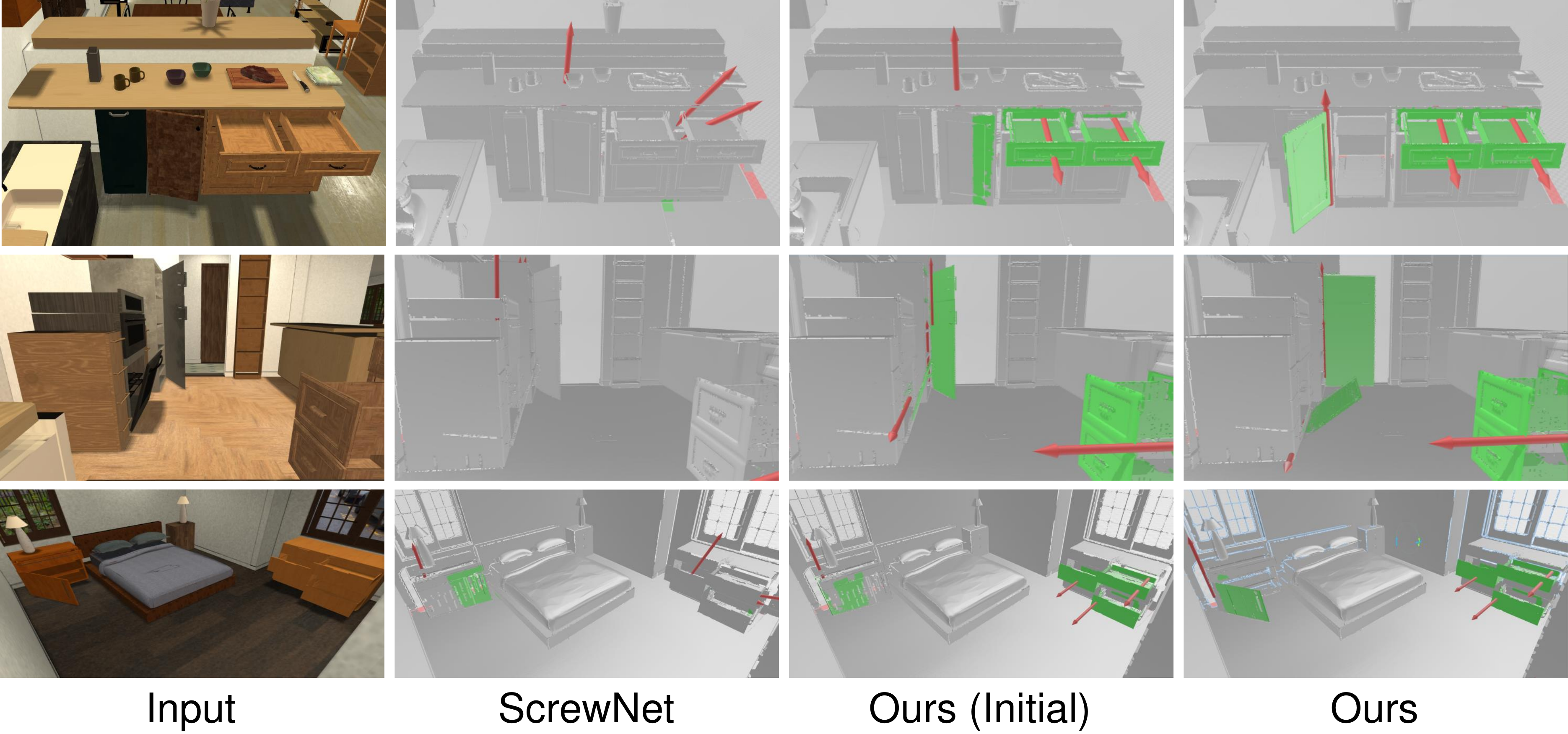}
    \vspace{-5mm}
    \caption{
        \textbf{Qualitative results on virtual scenes.} Static parts are colored grey, and mobile parts green. The estimated joints are shown as red arrows.
    }
    \label{fig:qual}
\end{figure}

\subsection{Experimental Setup}

\noindent  \textbf{Dataset.}
We train our model with data collected in the simulation environment. Specifically, we conduct experiments on the scenes provided by CubiCasa5K~\cite{kalervo2019cubicasa5k} dataset with the iGibson~\cite{li2021igibson} simulator. CubiCasa5K is a large-scale floorplan dataset covering over 5,000 scenes and 80 object categories, such as refrigerators and cabinets. We sample 2,500 scenes from the dataset and split them into (1,500, 500, 500) scenes for training, validation, and testing. We exclude objects with unopenable mobile parts due to collisions or other simulation issues. After the cleaning, our testing set covers 1,030 objects with 1,736 mobile parts.

\vspace{1mm}
\noindent  \textbf{Exploration-Driven Data Collection.}
We collect affordance supervision through robots' interactions. We first uniformly sample locations over the surface of both articulated and non-articulated parts, then have the robot interact with them. If the robot successfully moves any articulated part of the objects, we label the corresponding location as positive affordances or negative otherwise.
To simulate gripper-based interactions, we perform a collision check at each location to ensure enough space for placing a gripper. After that, we generate a pseudo-link between the object parts and the robot and then let the robot pull toward a certain direction, \ie, left, right, and backward. For simplification, we leverage a simplified virtual robot that can perform similar interactions as the real robot in a simulated environment. The virtual robot is a floating ball that makes single-point contact and can be teleported to any location in the scene. Such simplification lets us focus on the perception and interaction aspects without considering motion planning or navigation. 
For each successful interaction, we collect the robot's egocentric observations before and after interaction and the object's articulation model as training supervision.

\subsection{Training Details}
\noindent  \textbf{Affordance Network.}
We collect affordance data by the aforementioned exploration-driven method. Given the scene point cloud, we randomly sample points that fall on objects as potential interaction hotspots. By physically interacting with them, we label the potential interaction hotspots as \textit{positive} or \textit{negative} affordance depending on the interaction outcomes of \textit{success} versus \textit{failure}. The remaining points that have yet to be interacted with are ignored during training. The network takes the point clouds as input and is supervised under the affordance label. The data distribution is imbalanced due to the large proportion of negative data. To mitigate the imbalance problem, we optimize the network with the combination of the cross-entropy loss and the dice loss, as used in the previous work~\cite{deng20213d}.

\vspace{1mm}
\noindent  \textbf{Articulation Network.}
To collect object-centric observations before and after the interaction, the robot captures multi-view observations of the object at the front, right, and left sides of the object at a fixed distance if no collision occurs. After that, we aggregate them as a partial point cloud assuming the ground-truth camera poses are available. The network takes partial point clouds as input and is supervised by the ground truth articulation parameters provided by the simulator. We set up the loss functions and other training details following Ditto~\cite{jiang2022ditto}, except that the occupancy decoder is discarded.

\begin{table}[t]
    \scriptsize
    \centering
    \vspace{3mm}
    \caption{Quantitative results of affordance prediction.}
    \begin{tabular}{l|c|c|c}
    \toprule
\raisebox{\dimexpr-1\normalbaselineskip-2\cmidrulewidth-2\aboverulesep}[0pt][0pt]{Method} & \multicolumn{1}{c|}{Precision} & \multicolumn{2}{c}{Coverage} \\
      \cmidrule{2-4}
      &  & Prismatic & Revolute ($>15\degree$, $>30\degree$) \\
     \midrule
     3D AffordanceNet~\cite{deng20213d} & 0.56 & 0.23 & $(0.26,\, 0.03)$ \\
     Ours (w/o Refinement) & 0.66 & 0.60 & $(0.72,\, 0.10)$ \\
     Ours & - & 0.60 & $(0.72,\, 0.55)$ \\
     \bottomrule
    \end{tabular}
    \label{tab:affordance}
    \vspace{-5mm}
\end{table}

\subsection{Evaluation Metrics}
\noindent  \textbf{Affordance.}
To evaluate the affordance network, we report two metrics---coverage and precision. Coverage is the proportion of successfully interacted parts among the total number of interactable parts, and precision is the fraction of successful interactions that the agent attempted. For coverage of the revolute joint, we further define certain open degrees as the thresholds for successful interactions. Note that we do not define additional thresholds for prismatic joints since most prismatic joints are fully opened without refinements.

\vspace{1mm}
\noindent  \textbf{Articulation Model.}
To evaluate the estimated articulation model,~\ie, prismatic/revolute joint parameters and part segmentation, we use the same metrics as in Ditto~\cite{jiang2022ditto}. For both types of joints, we measure point-wise segmentation IoU of the mobile parts (Mobile Seg. IoU) and joint axis orientation error (Angle Err.). For the revolute joint, the position of the joint axis is also important, so we measure the position error (Pos Err.) by the distance between the predicted and ground-truth rotation axis.

\begin{table}[t]
    \scriptsize
    \centering
    \caption{Quantitative results of articulation inference.}
    \begin{tabular}{l|c|c|cc}
    \toprule
\raisebox{\dimexpr-1\normalbaselineskip-2\cmidrulewidth-2\aboverulesep}[0pt][0pt]{Method} & \multicolumn{1}{c|}{Geometry} & \multicolumn{3}{c}{Joint} \\
      \cmidrule{2-5}
      & \raisebox{\dimexpr-1\normalbaselineskip-1\cmidrulewidth-1\aboverulesep}[0pt][0pt] {\shortstack[c]{Mobile \\ Seg. IoU\,$\uparrow$}}  & Prismatic & \multicolumn{2}{c}{Revoulute} \\
      \cmidrule{3-5}
      & & Angle Err.\,$\downarrow$ & Angle Err.\,$\downarrow$ & Pos Err.\,$\downarrow$ \\
     \midrule
     ScrewNet~\cite{jain2020screwnet} & 0.34 & 0.28 & 46.19 & 0.13 \\
     Ours (w/o Refinement) & 0.78 & 0.04 & 49.0 & 0.06 \\
     Ours (w/o Regularity) & 0.78 & 0.05 & 31.0 & 0.05 \\
     Ours & 0.81 & 0.04 & 25.2 & 0.04 \\
     \bottomrule
    \end{tabular}
    \label{tab:articulation}
    \vspace{-5mm}
\end{table}

\subsection{Baselines}
\noindent \textbf{3D AffordanceNet~\cite{deng20213d}.}
To validate the efficacy of exploration-driven data collection, we compare the affordance labels collected from the robot's first-hand embodiment experience and manual affordance annotations. 3D AffordanceNet is a benchmark for visual object affordance understanding. It provides 3D affordance maps annotated by humans. We train an affordance model with the same architecture as ours using the ground-truth affordance map from 3D AffordanceNet and evaluate it on our test scenes.

\vspace{1mm}
\noindent \textbf{ScrewNet~\cite{jain2020screwnet}.}
To validate our joint parametrization choices, we modify our model's output and adopt the screw-based joint parametrization in ScrewNet~\cite{jain2020screwnet}. ScrewNet uses screw theory to unify the representation of different articulation types and perform category-independent articulation model estimation. While we follow Ditto~\cite{jiang2022ditto} and predict point-wise dense joints, ScrewNet predicts one global joint.

\vspace{1mm}
\noindent \textbf{Ours (w/o Refinement).}
We test an ablated version of our approach, where we only use the results from the initial stage.
Compared with our complete approach, Ours (w/o Refinement) does not refine the affordance based on observed articulated motion and no further refinement of the articulation model. This comparison helps validate the iterative refinement of affordance and articulation.
%

\vspace{1mm}
\noindent \textbf{Ours (w/o Regularity).}
We leverage action perception regularities by incorporating the contact regions of the interaction into our articulation network. In this ablated version of our approach, we remove the contact region information during training and inference. We present this result to validate the regularities for articulation inference.

\begin{figure*}[t]
    \centering
	\includegraphics[width=0.75\linewidth]{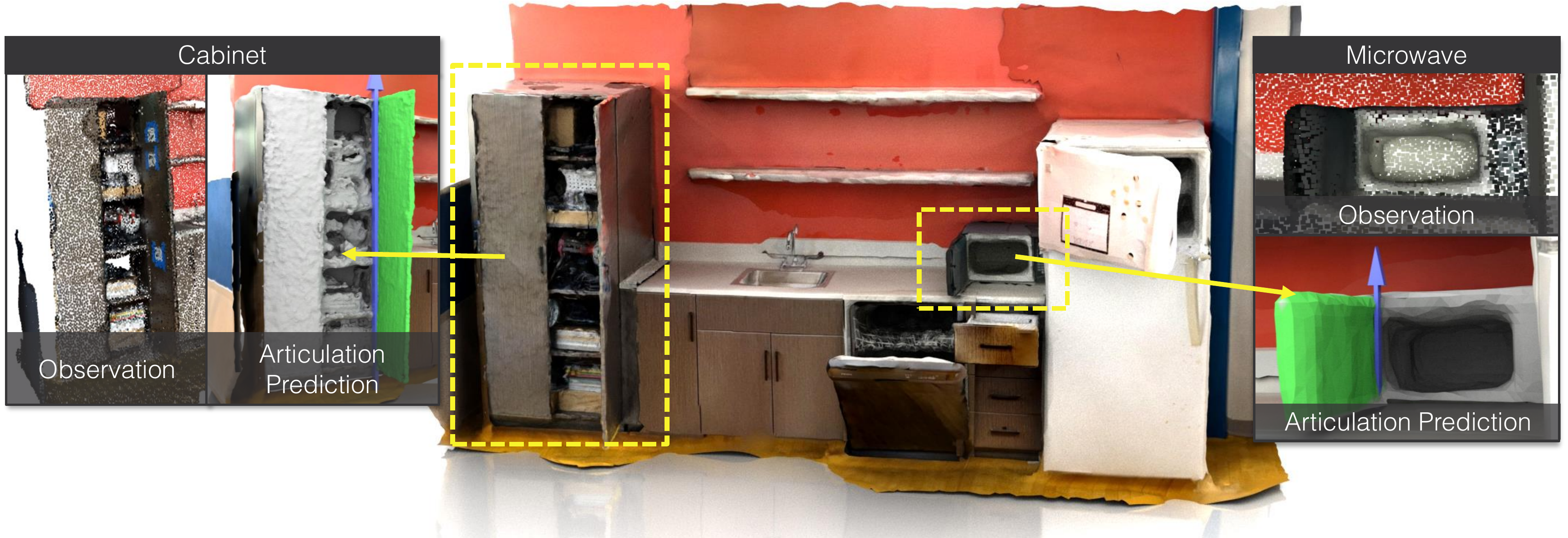}
    \vspace{-2mm}
    \caption{
        \textbf{Reconstructed articulation model of the real scene.}  Static parts are colored grey, and mobile parts are green. The estimated joints are visualized with blue arrows. More results on the same scene are shown in Figure~\ref{fig: teaser}.
    }
    \label{fig: realworld}
\end{figure*}

\subsection{Affordance Prediction}

The quantitative results of affordance prediction are shown in Table~\ref{tab:affordance}. Ours (w/o Refinement) obtains significantly better results than the model trained on 3D AffordanceNet. Human annotated affordance is biased towards human experience and might not generalize to robot manipulation. The distribution shift between the annotated scenes and the deployment environments also introduces extra challenges for generalization. In comparison, we train our affordance model with the data collected through self-exploration. The data is self-annotated based on the outcome of the robot's action executed in the environment. Therefore the trained model can easily adapt to the test scenes and shows much better interaction performance.

Our final model, with the iterative refinement of affordance and articulation, further boosts the coverage performance of revolute joints. The percentage of revolute joints opened over 30$^\circ$ increases from 10\% to 55\%. This improvement verifies the iterative refinement scheme. The initial estimation of the articulation model provides vital clues about the part-level affordance. The inferred part-level affordance creates more visible articulation motions and opens the parts with revolute joints to a larger degree. Note that our iterative refinement process does not add new or remove existing interaction hotspots. Therefore, we do not report the precision result after the refinement.

\subsection{Articulation Inference}

We show the results of articulation estimation in Table~\ref{tab:articulation}. ScrewNet-based model~\cite{jain2020screwnet} performs poorly on articulation estimation, especially the part segmentation. Ours (w/o Regularity) does not leverage the interaction information on where the action occurs and thus performs inferior to our complete model in part segmentation and joint axis prediction. In Figure~\ref{fig:qual}, we see that the ScrewNet-based model hardly segments any mobile parts. The joint axis predictions of this baseline are also far from correct. Ours (w/o Refinement) demonstrates much better results on object parts with prismatic joints. However, the result on the cabinet door with a revolute joint is less accurate. Ours (w/o Refinement) only segments less than half of the door, and the position of the estimated revolute joint axis also deviates from the correct location. Our refined model first opens the cabinet to a larger degree and reveals more previously occluded surfaces. With the new observation with more significant object state change, our refined model can predict more accurate part segmentation and joint parameters. 

\subsection{Real-World Evaluation}
Finally, we evaluate our method in a real-world household scene. We use the LiDAR and camera of an iPhone 12 Pro to recreate the scene in a 3D scan, rather than using a physical robot. We predict interaction hotspots and interact with the objects at these hotspots with our own hands. We then collect novel observations and run our approach to build the scene-level articulation model. The results in Figure~\ref{fig: teaser} and Figure~\ref{fig: realworld} show that our approach can be applied to the real scenario without any modification and reconstruct an accurate articulation model of the scene.

\subsection{Limitations}
We use the simulated robot grippers and human interactions to simplify the exploration and object manipulation process in virtual and real-world experiments. Thus, this work has the following limitations: a) \textit{Exploration:} To move between different locations, we directly teleport the robot. We abstract away the navigation and motion planning problems. Moreover, we assume perfect odometry and depth estimation while reconstructing the static model; b) \textit{Interaction:} We perform all interactions by creating pseudo links between the robot and objects or by humans. Issues such as joint constraints or self-collisions of the robot are not taken into consideration during object manipulation.








\section{Conclusion}
We develop an interactive perception approach to building scene-level articulation models. The robot physically interacts with articulated objects and infers their articulation properties from visual observations before and after exploratory actions. We further improve the quality of our prediction by coupling affordance prediction and articulation inference in an iterative procedure. Quantitative results demonstrate that our approach outperforms baselines by a substantial margin in both affordance prediction and articulation estimation. The ablation studies confirm that the iterative refinement process improves both tasks. Last, we demonstrate that our approach generalizes to real-world observations for creating an articulation model of a kitchen scene. These results manifest the promise of our approach in building interactive models for robot manipulation in everyday environments.

\vspace{2mm}
\noindent  \textbf{Acknowledgments} 
We would like to thank Huihan Liu, Mingyo Seo, and Soroush Nasiriany for providing feedback on this manuscript. This work has been partially supported by NSF CNS-1955523 and FRR-2145283, the MLL Research Award from the Machine Learning Laboratory at UT-Austin, and the Amazon Research Awards.



\renewcommand*{\bibfont}{\footnotesize}
\printbibliography 


\end{document}